\documentclass{article}
\usepackage{spconf,amsmath,graphicx,hyperref}
\usepackage{amssymb}
\usepackage{booktabs}
\usepackage{multirow} 
\usepackage{dblfloatfix}
\usepackage{cite}

\title{CBPNet: A Continual Backpropagation Prompt Network for Alleviating Plasticity Loss on Edge Devices}

\name{Runjie Shao$^{\star}$$^{\dagger}$ \qquad Boyu Diao\thanks{Corresponding author: diaoboyu2012@ict.ac.cn}$^{\star}$$^{\dagger}$ \qquad Zijia An$^{\star}$$^{\dagger}$ \qquad Ruiqi Liu$^{\star}$$^{\dagger}$ \qquad Yongjun Xu$^{\star}$$^{\dagger}$
}

\address{$^{\star}$ Institute of Computing Technology, Chinese Academy of Sciences, Beijing, China \\
$^{\dagger}$University of Chinese Academy of Sciences, Beijing, China
         }

\begin{document}

\ninept
\maketitle
\makeatletter
\renewcommand{\@makefnmark}{}
\makeatother
\footnotetext{© 2025 IEEE. Personal use of this material is permitted. Permission from IEEE must be obtained for all other uses, in any current or future media, including reprinting/republishing this material for advertising or promotional purposes, creating new collective works, for resale or redistribution to servers or lists, or reuse of any copyrighted component of this work in other works.}

\begin{abstract}
To meet the demands of applications like robotics and autonomous driving that require real-time responses to dynamic environments, efficient continual learning methods suitable for edge devices have attracted increasing attention. In this transition, using frozen pre-trained models with prompts has become a mainstream strategy to combat catastrophic forgetting. However, this approach introduces a new critical bottleneck: plasticity loss, where the model's ability to learn new knowledge diminishes due to the frozen backbone and the limited capacity of prompt parameters.
We argue that the reduction in plasticity stems from a lack of update vitality in underutilized parameters during the training process. To this end, we propose the Continual Backpropagation Prompt Network (CBPNet), an effective and parameter-efficient framework designed to restore the model's learning vitality. We innovatively integrate an Efficient CBP Block that counteracts plasticity decay by adaptively re-initializing these underutilized parameters.
Experimental results on edge devices demonstrate CBPNet’s effectiveness across multiple benchmarks. On Split CIFAR-100, it improves average accuracy by over 1\% against a strong baseline, and on the more challenging Split ImageNet-R, it achieves a state-of-the-art accuracy of 69.41\%. This is accomplished by training additional parameters that constitute less than 0.2\% of the backbone's size, validating our approach.

\end{abstract}
\begin{keywords}
Continual learning, Continual Backpropagation, Plasticity Loss, Edge Devices, Prompt-based Methods
\end{keywords}
\section{Introduction}
\label{sec:intro}

Continual learning (CL) enables models to learn sequentially from a continuous stream of data \cite{wang2024comprehensivesurveycontinuallearning}, a critical capability for applications on resource-constrained edge devices \cite{navardi2025genaiedgecomprehensivesurvey}. Applying pre-trained models (PTMs) to this challenge is a promising avenue, as their powerful prior knowledge provides a strong foundation against catastrophic forgetting \cite{1989Catastrophic}---the tendency to lose knowledge of previous tasks upon learning new ones \cite{ijcai2024p0924}. Prior approaches have largely relied on rehearsal-based methods \cite{buzzega2020darkexperiencegeneralcontinual,chaudhry2019tiny,8954008,10.1007/978-3-030-58536-5_31,9711466,an2025ior}, which are effective but memory-intensive, or regularization-based methods \cite{Kirkpatrick_2017,8107520}, which protect old knowledge but can overly constrain the model's plasticity. To this end, parameter-efficient prompt-based methods \cite{wang2022learningpromptcontinuallearning,wang2022dualpromptcomplementarypromptingrehearsalfree} have excelled by guiding a frozen PTM with a small set of trainable parameters, avoiding the need to store past data \cite{schwarz2018progress}. 

However, the limited capacity of the small number of trainable prompt parameters in these methods introduces a critical secondary problem: a loss of model plasticity \cite{petit2023plastil,Dohare2024Loss}. This issue manifests as a progressive decline in the model's capacity to absorb new knowledge, leading to a significant drop in performance during later learning stages. The challenge is particularly pronounced on resource-constrained edge devices \cite{asadi2014survey,liu2024continual,liu2025low}, where compensating for this loss by training additional parameters or storing extensive historical data is infeasible \cite{xu2021privacypreservingmachinelearningmethods}.

To address this challenge, this paper introduces the Continual Backpropagation Prompt Network (CBPNet). Inspired by the recent findings of Dohare et al.  \cite{Dohare2024Loss}, which demonstrate that reinitializing "stagnant" neurons alleviates plasticity loss, we argue this principle is critical for prompt-based methods where the vitality of few trainable parameters is fundamental for adaptation. We therefore embed a modular Continual Backpropagation (CBP) mechanism into the DualPrompt \cite{wang2022dualpromptcomplementarypromptingrehearsalfree} framework. This mechanism monitors and reinitializes neurons with minimal contributions, injecting vitality for new tasks without compromising prior knowledge and significantly enhancing learning capabilities with minimal overhead.

Our contributions are threefold: 1) We propose CBPNet, a framework that mitigates plasticity loss in prompt-based continual learning by integrating a modular CBP mechanism. 2) Its core is a modular block with a bottleneck architecture that safely revitalizes the network by re-initializing low-utility neurons. 3) We demonstrate state-of-the-art performance on Split CIFAR-100 and ImageNet-R with a trainable overhead of less than 0.2\% of the backbone.

\section{Methodology}
\label{sec:pagestyle}

In this section, we elaborate on the architecture and key components of the proposed CBPNet (Continual Backpropagation Prompt Network). CBPNet is designed to integrate DualPrompt\cite{wang2022dualpromptcomplementarypromptingrehearsalfree}’s efficient prompt learning mechanism with the plasticity-loss mitigation capability of Continuous Backpropagation (CBP), enabling efficient on-device continual learning.

\subsection{The overall architecture of CBPNet}
Similar to other works\cite{wang2022learningpromptcontinuallearning,wang2022dualpromptcomplementarypromptingrehearsalfree}, we employ a Vision Transformer (ViT)\cite{dosovitskiy2021imageworth16x16words} pre-trained on large-scale datasets (e.g., ImageNet-21k) as the feature extractor. During continual learning, most parameters of the ViT backbone remain frozen to reduce computational and storage costs while preserving general visual knowledge. 
The overall architecture of CBPNet is shown in Figure \ref{fig:CBDNET}. It is primarily composed of the following components:

\begin{figure*}[htb] 
\centering 
\includegraphics[width=\textwidth]{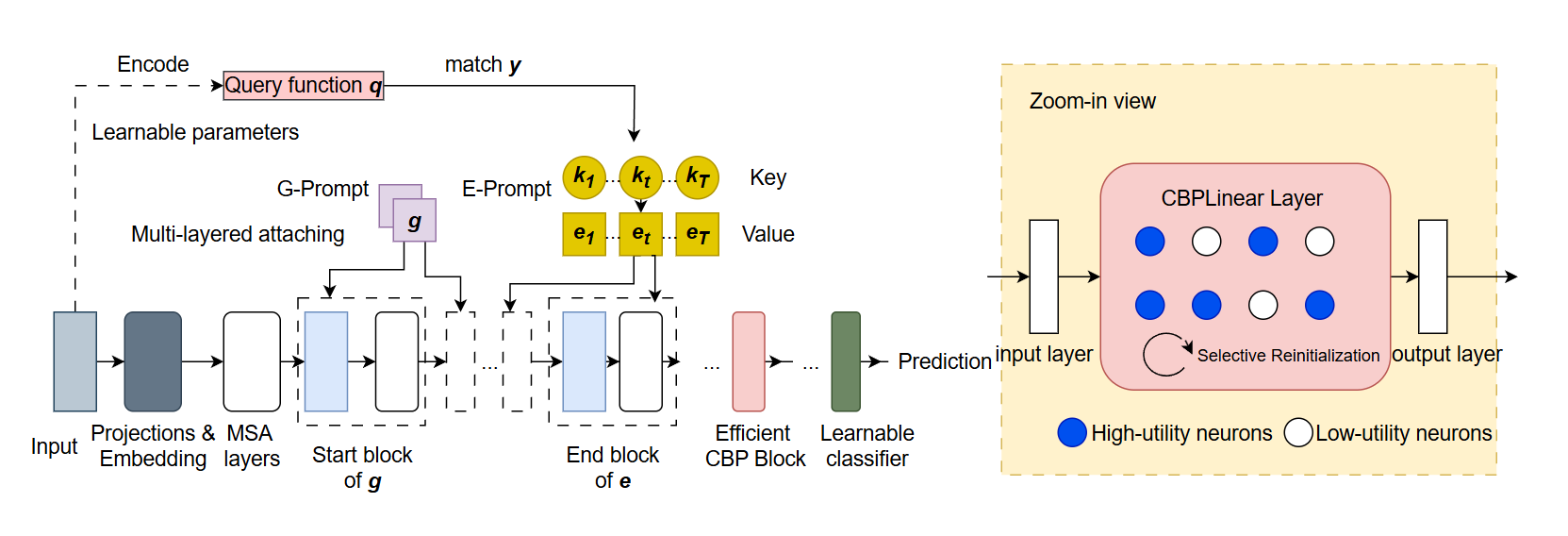} 
\caption{The overall architecture of our proposed CBPNet. To address plasticity loss, we introduce an Efficient CBP Block after the prompt-guided ViT backbone. The zoom-in view illustrates the selective re-initialization of low-utility neurons within the Efficient CBP Block.}
\label{fig:CBDNET}
\end{figure*}

\subsection{DualPrompt Module}
We adopt DualPrompt \cite{wang2022dualpromptcomplementarypromptingrehearsalfree} for continual learning on a frozen ViT backbone, introducing lightweight, learnable prompts to acquire new knowledge while mitigating forgetting. Specifically, a task-shared G-Prompt $G \in \mathbb{R}^{L_G \times D}$ is inserted into shallow Transformer layers to capture common knowledge, while task-specific E-Prompts $e_t \in \mathbb{R}^{L_E \times D}$ from a pool $E=\{e_t\}_{t=1}^T$ are placed in deeper layers. Each E-Prompt is associated with a learnable key $k_t \in \mathbb{R}^D$, enabling the model to select the most relevant prompt during inference via cosine similarity with the input's query vector. Prompts are integrated using Prefix-Tuning \cite{li2021prefix}, which prepends prompt-derived vectors $p_K$ and $p_V$ to the Key and Value matrices in the attention mechanism, thus influencing the model's focus without altering the input sequence length:
\begin{equation}
    f_{\text{prompt}}^{\text{Pre-T}}(p,h) = \text{MSA}(h_Q, [p_K; h_K], [p_V; h_V])
\end{equation}
The training objective jointly optimizes $(G, e_t, k_t)$ and a classification head $\phi$ by minimizing a combined loss of standard classification loss $\mathcal{L}$ (e.g., cross-entropy) and matching loss $\mathcal{L}_{match}$, which maximizes the similarity between an input and its corresponding task key:
\begin{equation}
    \min_{g,e_t,k_t,\phi} \mathcal{L}\big(f_{\phi}(f_{g,e_t}(x)), y\big) + \lambda\,\mathcal{L}_{match}(x,k_t)
\end{equation}
where $\lambda$ is a balancing hyperparameter.

\subsection{Efficient CBP Block}
CBPNet introduces an Efficient Continuous Backpropagation (CBP) Block positioned strategically after the ViT feature extractor (specifically, following the Global Pool layer after Transformer\cite{vaswani2017attention} blocks integrated with G-Prompt/E-Prompt) and before the final classification head. This innovative block is designed to combat the decline in plasticity that occurs when the ViT backbone is largely frozen during the learning of new tasks\cite{zhang2023adapter}. The core of the CBP mechanism lies in its ability to maintain the network's capacity to acquire new knowledge by monitoring the "utility"\cite{Dohare2024Loss} of internal units (neurons/weights within the Efficient CBP Block's linear layers) and selectively re-initializing those with low utility. The Efficient CBP Block's internal structure comprises an Input Layer, a GELU Activation function, one CBP Linear Layer where the continuous backpropagation mechanism operates, and an Output Layer.

We implement CBP as a modular, post-hoc block rather than altering the backbone training, a deliberate design that decouples the mechanism from the pre-trained model. This ensures safety by isolating dynamic re-initialization to preserve the frozen backbone’s knowledge, and efficiency by leveraging an internal bottleneck structure that maintains plasticity with minimal parameter overhead, making it well-suited for continual learning under device constraints.

\subsubsection{Contribution Utility}
We use contribution utility $u_{i}^{(l)}$ to quantify the importance of the $i$-th hidden unit in the $l$-th core linear layer of the Efficient CBP Block. The motivation is to identify stagnant or less-used units that have ceased to contribute meaningfully to the network's function. These low-utility units can then be selectively re-initialized to restore the network's plasticity. The utility is defined as:
\begin{equation}\label{eq6}
u_{i}^{(l)} = \eta \times u_{i}^{(l)} + (1 - \eta) \times |\mathbf{h}_{l,i,t}| \times \left| \sum_{k=1}^{n_{l+1}} \mathbf{w}_{l,i,k,t} \right|
\end{equation}
where $\mathbf{h}_{l,i,t}$ is the $i$-th unit's output at time $t$, $\mathbf{w}_{l,i,k,t}$ is the weight from unit $i$ to unit $k$ in the next layer, and $\eta$ controls historical utility retention. This metric is effective because it holistically evaluates a unit's importance by combining its own activation magnitude ($|\mathbf{h}_{l,i,t}|$) with its collective influence on all units in the subsequent layer ($\left| \sum \mathbf{w}_{l,i,k,t} \right|$). The use of an exponential moving average ensures this assessment is stable, reflecting a unit's long-term contribution rather than fluctuating with individual data batches.

\subsubsection{Selective Re-initialization Strategy}
The Efficient CBP Block implements selective re-initialization via:  
1. Age tracking: Each unit has an age counter $a_{i}^{(l)}$ that increments with updates. Only units exceeding a maturity threshold $m$ are candidates for re-initialization. This mechanism acts as a crucial "grace period," as newly re-initialized units start with zero utility due to their zeroed-out output weights. The maturity threshold prevents these new units from being immediately selected for re-initialization again, allowing them sufficient time to participate in training and develop a meaningful utility score. 
2. Utility evaluation: For mature units, those with utility $u_{i}^{(l)}$ below a threshold (or low-ranked) are re-initialized with probability $\rho$.  

Re-initialization resamples the unit's input weights from the initial distribution and zeros its output weights, minimizing the impact on prior knowledge while enabling new learning.  

After each training batch, mature units' utilities are evaluated, and low-utility ones are randomly re-initialized at a rate $\rho$.

\subsection{Training and  Inference Procedure.}
During the training phase, CBPNet freezes the parameters of the ViT backbone network and focuses on optimizing only the lightweight components: the G-Prompt, E-Prompts and their keys, the newly added Efficient CBP Block, and a task-specific classification head. The model adopts the classification loss and matching loss from the DualPrompt\cite{wang2022dualpromptcomplementarypromptingrehearsalfree} framework. The core CBP re-initialization mechanism is automatically activated during the backpropagation and parameter update processes that occur within its dedicated block. In the inference phase, an input is processed sequentially: it first passes through the shallow layers of ViT with the G-Prompt, then the deep layers with the selected E-Prompt, followed by the Efficient CBP Block, and finally outputs the prediction results through the classification head.

\subsection{Efficiency and Complexity Analysis}
The described procedures ensure CBPNet is a highly efficient method suitable for on-device learning. In terms of space complexity, the additional parameters are minimal. Let $N$ be the parameters of the PTM backbone; the number of new trainable parameters from the prompts ($P_{prompts}$) and our CBP block ($P_{cbp}$) is significantly smaller, such that $P_{prompts} + P_{cbp} \ll N$. Furthermore, as a rehearsal-free method, CBPNet requires no extra memory for storing past data, a major advantage over buffer-based approaches. Regarding time complexity, the computation during training is dominated by the forward and backward propagation through the large ViT backbone. The overhead introduced by our Efficient CBP Block, which consists of a small MLP and an efficient utility calculation, is marginal. Crucially for on-device deployment, the re-initialization mechanism is inactive during inference. The CBP block then functions as a simple feed-forward pass, resulting in an inference time that is nearly identical to that of the standard DualPrompt \cite{wang2022dualpromptcomplementarypromptingrehearsalfree} baseline.

\section{Experiment}
\label{sec:typestyle}
\subsection{Datasets and Implementation Details}

We evaluated CBPNet's performance on two challenging class-incremental learning benchmarks: Split CIFAR-100 \cite{2009Learning} and Split ImageNet-R \cite{hendrycks2021many}. For both benchmarks, we adopted a class-incremental setting with 10 sequential tasks. The backbone network was a pre-trained ViT-B/16, and input images were resized to 224x224. We compared CBPNet against various continual learning methods, including ER \cite{chaudhry2019tiny}, DER++ \cite{buzzega2020darkexperiencegeneralcontinual}, GDumb \cite{10.1007/978-3-030-58536-5_31}, EWC \cite{Kirkpatrick_2017}, LwF \cite{8107520}, L2P \cite{wang2022learningpromptcontinuallearning}, DualPrompt\cite{wang2022dualpromptcomplementarypromptingrehearsalfree} (our direct baseline), FT-seq, BiC\cite{8954008}, and Co$^2$L \cite{9711466}, with the joint-training upper-bound as a reference. For CBPNet, the G-Prompt length was 5 (applied to the first 2 Transformer layers , 0 and 1), and the E-Prompt length was 5 (applied to Transformer layers 2 through 4), using Prefix-Tuning. The maturity threshold of Efficient CBP Block $m$ was set to 1000, and the replacement rate $\rho$ to $10^{-5}$. Training employed the Adam optimizer, with learning rate and batch size referenced from DualPrompt\cite{wang2022dualpromptcomplementarypromptingrehearsalfree} and adjusted based on the memory constraints of the NVIDIA Jetson Orin Nano; the total training epochs were consistent with baseline methods for a fair comparison. The evaluation metrics were the average accuracy and forgetting rate after learning all tasks.

\subsection{ Baselines and Evaluation Metrics}

To provide a comprehensive evaluation, we compare CBPNet against three categories of representative continual learning methods. 

\textbf{Rehearsal-based methods}, such as ER \cite{chaudhry2019tiny} and DER++ \cite{buzzega2020darkexperiencegeneralcontinual}, BiC \cite{8954008}, GDumb \cite{10.1007/978-3-030-58536-5_31}, Co$^2$L \cite{9711466}, which store past data. 

\textbf{Regularization-based methods}, including EWC \cite{Kirkpatrick_2017} and LwF \cite{8107520}, adds loss terms to penalize changes to important weights, thus preserving old knowledge. 

\textbf{Prompt-based methods}, such as L2P\cite{wang2022learningpromptcontinuallearning} and our direct baseline DualPrompt\cite{wang2022dualpromptcomplementarypromptingrehearsalfree}, keep the backbone frozen and only tune small prompt parameters. 

We also include sequential fine-tuning (FT-seq) as a lower bound for performance and joint training on all data (Upper-bound)  as a practical performance ceiling.

We evaluate all methods using two standard metrics to measure their continual learning capabilities.
Average Accuracy is calculated as the average of the test accuracies over all tasks after the entire training sequence is complete, reflecting the overall performance of the model. Forgetting Rate measures the average decrease in accuracy on previously learned tasks after the model has been trained on new, subsequent tasks, directly quantifying the degree of catastrophic forgetting.

\begin{table*}[t]
\centering
\caption{Performance comparison on Split CIFAR-100 and Split ImageNet-R benchmarks.}
\label{tab:full_results}
\resizebox{\textwidth}{!}{
\begin{tabular}{l||c|cc||c|cc}
\hline
\multirow{2}{*}{\textbf{Method}} & \multirow{2}{*}{\textbf{Buffer size}} & \multicolumn{2}{c||}{\textbf{Split CIFAR-100}} & \multirow{2}{*}{\textbf{Buffer size}} & \multicolumn{2}{c}{\textbf{Split ImageNet-R}} \\ \cline{3-4} \cline{6-7}
 & & \textbf{Avg. Acc (↑)} & \textbf{Forgetting (↓)} & & \textbf{Avg. Acc (↑)} & \textbf{Forgetting (↓)} \\ \hline \hline
ER \cite{chaudhry2019tiny} & \multirow{5}{*}{1000} & 67.43 $\pm$ 0.57 & 33.05 $\pm$ 1.27 & \multirow{5}{*}{1000} & 54.81 $\pm$ 1.28 & 35.12 $\pm$ 0.52 \\
BiC \cite{8954008} & & 65.62 $\pm$ 1.75 & 34.93 $\pm$ 1.63 & & 51.98 $\pm$ 1.08 & 36.57 $\pm$ 1.05 \\
GDumb \cite{10.1007/978-3-030-58536-5_31} & & 66.52 $\pm$ 0.37 & -- & & 38.15 $\pm$ 0.55 & -- \\
DER++ \cite{buzzega2020darkexperiencegeneralcontinual} & & 60.59 $\pm$ 0.86 & 39.73 $\pm$ 0.99 & & 55.03 $\pm$ 1.30 & 34.50 $\pm$ 1.49 \\
Co$^2$L \cite{9711466} & & 71.58 $\pm$ 1.31 & 28.32 $\pm$ 1.55 & & 52.97 $\pm$ 1.54 & 37.12 $\pm$ 1.80 \\
\hline
ER \cite{chaudhry2019tiny} & \multirow{5}{*}{5000} & 81.82 $\pm$ 0.17 & 16.35 $\pm$ 0.25 & \multirow{5}{*}{5000} & 64.92 $\pm$ 0.40 & 23.11 $\pm$ 0.88 \\
BiC \cite{8954008} & & 81.20 $\pm$ 0.85 & 17.22 $\pm$ 1.02 & & 64.12 $\pm$ 1.26 & 22.09 $\pm$ 1.72 \\
GDumb \cite{10.1007/978-3-030-58536-5_31} & & 81.16 $\pm$ 0.02 & -- & & 65.57 $\pm$ 0.28 & -- \\
DER++ \cite{buzzega2020darkexperiencegeneralcontinual} & & 83.33 $\pm$ 0.34 & 14.43 $\pm$ 0.72 & & 66.21 $\pm$ 0.86 & 20.59 $\pm$ 1.23 \\
Co$^2$L \cite{9711466} & & 82.23 $\pm$ 0.89 & 17.34 $\pm$ 1.79 & & 65.37 $\pm$ 0.14 & 23.21 $\pm$ 0.71 \\
\hline
FT-seq & \multirow{6}{*}{0} & 33.41 $\pm$ 0.84 & 86.15 $\pm$ 0.20 & \multirow{6}{*}{0} & 28.64 $\pm$ 1.35 & 63.61 $\pm$ 1.49 \\
EWC \cite{Kirkpatrick_2017} & & 46.73 $\pm$ 0.29 & 32.97 $\pm$ 1.16 & & 34.82 $\pm$ 0.43 & 55.88 $\pm$ 0.87 \\
LwF \cite{8107520} & & 60.10 $\pm$ 0.62 & 27.65 $\pm$ 2.16 & & 38.40 $\pm$ 1.22 & 52.01 $\pm$ 0.64 \\
L2P \cite{wang2022learningpromptcontinuallearning} & & 83.33 $\pm$ 0.28 & 7.29 $\pm$ 0.38 & & 61.26 $\pm$ 0.66 & 9.68 $\pm$ 0.47 \\
DualPrompt \cite{wang2022dualpromptcomplementarypromptingrehearsalfree}& & 85.30 $\pm$ 0.33 & 5.16 $\pm$ 0.09 & & 67.74 $\pm$ 0.51 & 4.65 $\pm$ 0.21 \\
\textbf{CBPNet (Ours)} & & \textbf{86.31 $\pm$ 0.27} & \textbf{5.07 $\pm$ 0.07} & & \textbf{69.41 $\pm$ 0.42} & \textbf{4.45 $\pm$ 0.16} \\ \hline
Upper-bound & -- & 90.85 $\pm$ 0.12 & -- & -- & 79.13 $\pm$ 0.18 & -- \\ \hline
\end{tabular}}
\end{table*}

\begin{table}[ht]
\centering
\caption{Ablation study. }
\label{tab:ablation_combined}
\resizebox{0.48\textwidth}{!}{%
\begin{tabular}{lcc|cc}
\hline
\multirow{2}{*}{\textbf{Model Variant}} & \multirow{2}{*}{\textbf{DualPrompt\cite{wang2022dualpromptcomplementarypromptingrehearsalfree}}} & \multirow{2}{*}{\textbf{Efficient CBP Block}} & \multicolumn{2}{c}{\textbf{Accuracy (\%)}} \\ \cline{4-5}
 & & & \textbf{CIFAR-100} & \textbf{ImageNet-R} \\
\hline
FT-seq & $\times$ & $\times$ & 33.41 & 28.64 \\
FT-seq + CBP & $\times$ & $\checkmark$ & 41.29 & 32.71 \\
DualPrompt\cite{wang2022dualpromptcomplementarypromptingrehearsalfree} & $\checkmark$ & $\times$ & 85.30 & 68.13 \\
\textbf{CBPNet (Ours)} & $\checkmark$ & $\checkmark$ & \textbf{86.31} & \textbf{69.41} \\
\hline
\end{tabular}
}
\end{table}

\begin{figure}[t]

\begin{minipage}[b]{.48\linewidth}
  \centering
  \centerline{\includegraphics[width=4.0cm]{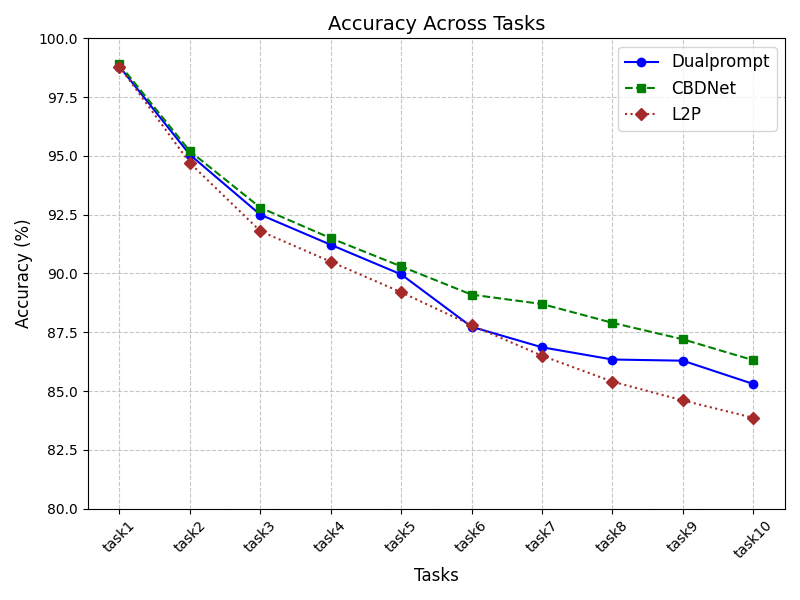}}
  \centerline{(a) Split CIFAR-100}\medskip
\end{minipage}
\hfill
\begin{minipage}[b]{0.48\linewidth}
  \centering
  \centerline{\includegraphics[width=4.0cm]{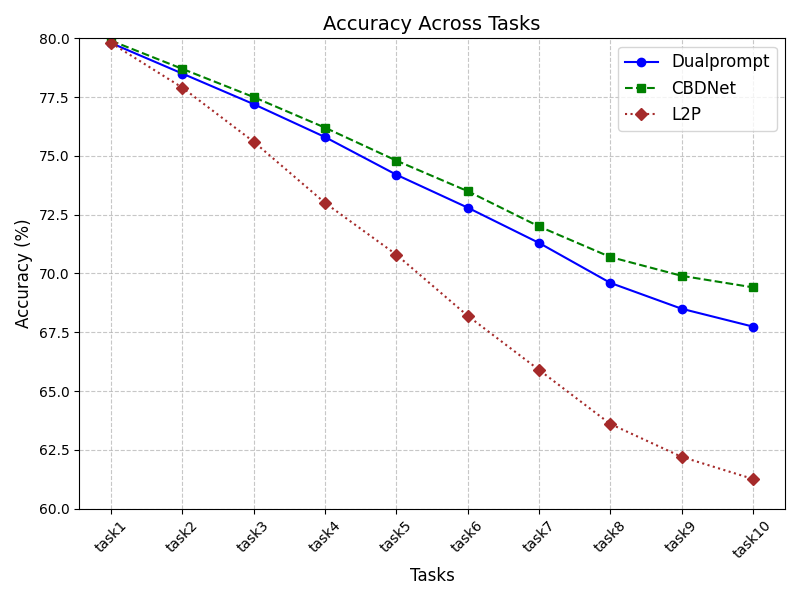}}
  \centerline{(b) Split ImageNet-R}\medskip
\end{minipage}
\caption{The accuracy performance on different datasets.}
\label{fig:res}
\end{figure}

\subsection{Comparison with State-of-the-Art}

Table \ref{tab:full_results} shows the performance comparison of CBPNet with other continual learning methods. Under the rehearsal-free setting (Buffer size 0), CBPNet achieves a state-of-the-art average accuracy of 86.31\% on Split CIFAR-100 and 69.41\% on Split ImageNet-R. On both datasets, our method improves upon the strong DualPrompt\cite{wang2022dualpromptcomplementarypromptingrehearsalfree} baseline, with the advantage being more pronounced on the more challenging ImageNet-R benchmark. Notably, CBPNet also demonstrates superior performance compared to other rehearsal-free methods like L2P\cite{wang2022learningpromptcontinuallearning}. Crucially for edge devices, CBPNet remains competitive with large-buffer (5000 samples) rehearsal-based methods, despite requiring no extra data storage. This indicates that by introducing the CBP mechanism, CBPNet effectively enhances the model's plasticity.

\subsection{Ablation Study}
The ablation results in Table \ref{tab:ablation_combined} validate our design, proving that both DualPrompt\cite{wang2022dualpromptcomplementarypromptingrehearsalfree} and our Efficient CBP Block play indispensable and synergistic roles. On Split CIFAR-100, applying DualPrompt\cite{wang2022dualpromptcomplementarypromptingrehearsalfree} alone to a fine-tuning baseline (FT-seq) significantly increases accuracy from 33.41\% to 85.30\%, forming a solid foundation against catastrophic forgetting. Building upon this, the introduction of our Efficient CBP Block (CBPNet) further improves the accuracy to 86.31\%. This trend is consistent on Split ImageNet-R, where CBPNet (69.41\%) provides a clear gain over the DualPrompt\cite{wang2022dualpromptcomplementarypromptingrehearsalfree} baseline (68.13\%). It is noteworthy that using the CBP block alone (FT-seq + CBP) also improves performance over the FT-seq , confirming its direct contribution. This demonstrates the standalone effectiveness of our proposed block.

\subsection{Analysis of Plasticity and Efficiency}
As shown in Figure \ref{fig:res}, we plotted the accuracy curves of different methods during the learning process. Traditional FT-seq (sequential fine-tuning) (not shown, but its accuracy drops sharply) and LwF\cite{8107520} exhibit severe performance degradation. DualPrompt\cite{wang2022dualpromptcomplementarypromptingrehearsalfree}'s performance is relatively stable, but its accuracy curve still shows a noticeable downward trend in the later stages. In contrast, CBPNet's accuracy curve (green squares) is the most stable throughout the entire learning process. Especially in the last few tasks, its performance drop is significantly smaller than that of DualPrompt\cite{wang2022dualpromptcomplementarypromptingrehearsalfree}, which intuitively demonstrates the advantage of CBPNet in mitigating plasticity loss.
This resilience is particularly evident in the later stages of training on Split ImageNet-R in Figure \ref{fig:res}.b. For instance, from task 6 to 10, the accuracy of DualPrompt\cite{wang2022dualpromptcomplementarypromptingrehearsalfree} declines by approximately 3.5\%, whereas CBPNet's accuracy drops by less than 2\%. This demonstrates that as task sequences lengthen and plasticity becomes more critical, the benefits of our Efficient CBP Block become increasingly pronounced.

In terms of efficiency, CBPNet inherits DualPrompt \cite{wang2022dualpromptcomplementarypromptingrehearsalfree}'s parameter-efficient philosophy by only training a small number of parameters while keeping the vast ViT-B/16 backbone (approximately 86 million parameters) frozen. This is achieved by complementing the prompts with our novel Efficient CBP block, a lightweight MLP designed with a bottleneck architecture. In total, the approximately 150,000 trainable parameters in CBPNet represent less than 0.2\% of the backbone's size. This highly efficient design drastically reduces both storage and computational requirements compared to full fine-tuning, making CBPNet a practical solution for maintaining plasticity on resource-constrained edge devices. Furthermore, in our experiments on a Jetson Orin (4GB), we observed that even when a reduced batch size led to a slight drop in accuracy, CBPNet's performance consistently surpassed that of other prompt-based methods, validating its real-world applicability under such constraints.

\section{Conclusions}
This paper proposed CBPNet, an efficient method that addresses plasticity loss in prompt-based learning by integrating an Efficient CBP Block with the DualPrompt\cite{wang2022dualpromptcomplementarypromptingrehearsalfree} framework to successfully maintain the model's ability to learn new tasks. Experiments on both Split CIFAR-100 and the more challenging Split ImageNet-R demonstrate that CBPNet significantly improves average accuracy over strong baselines, all while training a parameter set that is less than 0.2\% of the frozen backbone. This method provides a simple yet effective solution for deploying high-performance and adaptive AI systems on resource-constrained edge devices. Future work could explore applying our modular CBP mechanism to other domains, such as natural language processing, or combining it with memory-based strategies to achieve an even better stability-plasticity balance.


\vfill\pagebreak

\bibliographystyle{IEEEbib}
\bibliography{strings,refs}

\begin{thebibliography}{10}

\bibitem{wang2024comprehensivesurveycontinuallearning}
Liyuan Wang, Xingxing Zhang, Hang Su, and Jun Zhu,
\newblock ``A comprehensive survey of continual learning: Theory, method and application,'' 2024.

\bibitem{navardi2025genaiedgecomprehensivesurvey}
Mozhgan Navardi, Romina Aalishah, Yuzhe Fu, Yueqian Lin, Hai Li, Yiran Chen, and Tinoosh Mohsenin,
\newblock ``Genai at the edge: Comprehensive survey on empowering edge devices,'' 2025.

\bibitem{1989Catastrophic}
Michael Mccloskey and Neal~J. Cohen,
\newblock ``Catastrophic interference in connectionist networks: The sequential learning problem,''
\newblock {\em Psychology of Learning and Motivation}, vol. 24, pp. 109--165, 1989.

\bibitem{ijcai2024p0924}
Da-Wei Zhou, Hai-Long Sun, Jingyi Ning, Han-Jia Ye, and De-Chuan Zhan,
\newblock ``Continual learning with pre-trained models: A survey,''
\newblock in {\em Proceedings of the Thirty-Third International Joint Conference on Artificial Intelligence, {IJCAI-24}}, Kate Larson, Ed. 8 2024, pp. 8363--8371, International Joint Conferences on Artificial Intelligence Organization,
\newblock Survey Track.

\bibitem{buzzega2020darkexperiencegeneralcontinual}
Pietro Buzzega, Matteo Boschini, Angelo Porrello, Davide Abati, and Simone Calderara,
\newblock ``Dark experience for general continual learning: a strong, simple baseline,'' 2020.

\bibitem{chaudhry2019tiny}
Arslan Chaudhry, Marcus Rohrbach, Mohamed Elhoseiny, Thalaiyasingam Ajanthan, Puneet~K Dokania, Philip~HS Torr, and Marc'Aurelio Ranzato,
\newblock ``On tiny episodic memories in continual learning,''
\newblock {\em arXiv preprint arXiv:1902.10486}, 2019.

\bibitem{8954008}
Yue Wu, Yinpeng Chen, Lijuan Wang, Yuancheng Ye, Zicheng Liu, Yandong Guo, and Yun Fu,
\newblock ``Large scale incremental learning,''
\newblock in {\em 2019 IEEE/CVF Conference on Computer Vision and Pattern Recognition (CVPR)}, 2019, pp. 374--382.

\bibitem{10.1007/978-3-030-58536-5_31}
Ameya Prabhu, Philip H.~S. Torr, and Puneet~K. Dokania,
\newblock ``Gdumb: A simple approach that questions our progress in continual learning,''
\newblock in {\em Computer Vision -- ECCV 2020}, Andrea Vedaldi, Horst Bischof, Thomas Brox, and Jan-Michael Frahm, Eds., Cham, 2020, pp. 524--540, Springer International Publishing.

\bibitem{9711466}
Hyuntak Cha, Jaeho Lee, and Jinwoo Shin,
\newblock ``Co2l: Contrastive continual learning,''
\newblock in {\em 2021 IEEE/CVF International Conference on Computer Vision (ICCV)}, 2021, pp. 9496--9505.

\bibitem{an2025ior}
Zijia An, Boyu Diao, Libo Huang, Ruiqi Liu, Zhulin An, and Yongjun Xu,
\newblock ``Ior: Inversed objects replay for incremental object detection,''
\newblock in {\em ICASSP 2025-2025 IEEE International Conference on Acoustics, Speech and Signal Processing (ICASSP)}. IEEE, 2025, pp. 1--5.

\bibitem{Kirkpatrick_2017}
James Kirkpatrick, Razvan Pascanu, Neil Rabinowitz, Joel Veness, Guillaume Desjardins, Andrei~A. Rusu, Kieran Milan, John Quan, Tiago Ramalho, Agnieszka Grabska-Barwinska, Demis Hassabis, Claudia Clopath, Dharshan Kumaran, and Raia Hadsell,
\newblock ``Overcoming catastrophic forgetting in neural networks,''
\newblock {\em Proceedings of the National Academy of Sciences}, vol. 114, no. 13, pp. 3521–3526, Mar. 2017.

\bibitem{8107520}
Zhizhong Li and Derek Hoiem,
\newblock ``Learning without forgetting,''
\newblock {\em IEEE Transactions on Pattern Analysis and Machine Intelligence}, vol. 40, no. 12, pp. 2935--2947, 2018.

\bibitem{wang2022learningpromptcontinuallearning}
Zifeng Wang, Zizhao Zhang, Chen-Yu Lee, Han Zhang, Ruoxi Sun, Xiaoqi Ren, Guolong Su, Vincent Perot, Jennifer Dy, and Tomas Pfister,
\newblock ``Learning to prompt for continual learning,'' 2022.

\bibitem{wang2022dualpromptcomplementarypromptingrehearsalfree}
Zifeng Wang, Zizhao Zhang, Sayna Ebrahimi, Ruoxi Sun, Han Zhang, Chen-Yu Lee, Xiaoqi Ren, Guolong Su, Vincent Perot, Jennifer Dy, and Tomas Pfister,
\newblock ``Dualprompt: Complementary prompting for rehearsal-free continual learning,'' 2022.

\bibitem{schwarz2018progress}
Jonathan Schwarz, Wojciech Czarnecki, Jelena Luketina, Agnieszka Grabska-Barwinska, Yee~Whye Teh, Razvan Pascanu, and Raia Hadsell,
\newblock ``Progress \& compress: A scalable framework for continual learning,''
\newblock in {\em International conference on machine learning}. PMLR, 2018, pp. 4528--4537.

\bibitem{petit2023plastil}
Gregoire Petit, Adrian Popescu, Eden Belouadah, David Picard, and Bertrand Delezoide,
\newblock ``Plastil: Plastic and stable exemplar-free class-incremental learning,''
\newblock in {\em Conference on lifelong learning agents}. PMLR, 2023, pp. 399--414.

\bibitem{Dohare2024Loss}
Shibhansh Dohare, J.~Fernando Hernandez-Garcia, Qingfeng Lan, Parash Rahman, A.~Rupam Mahmood, and Richard~S. Sutton,
\newblock ``Loss of plasticity in deep continual learning,''
\newblock {\em Nature}, vol. 632, pp. 768--774, 2024.

\bibitem{asadi2014survey}
Arash Asadi, Qing Wang, and Vincenzo Mancuso,
\newblock ``A survey on device-to-device communication in cellular networks,''
\newblock {\em IEEE Communications Surveys \& Tutorials}, vol. 16, no. 4, pp. 1801--1819, 2014.

\bibitem{liu2024continual}
Ruiqi Liu, Boyu Diao, Libo Huang, Zijia An, Zhulin An, and Yongjun Xu,
\newblock ``Continual learning in the frequency domain,''
\newblock {\em Advances in Neural Information Processing Systems}, vol. 37, pp. 85389--85411, 2024.

\bibitem{liu2025low}
Ruiqi Liu, Boyu Diao, Libo Huang, Zijia An, Hangda Liu, Zhulin An, and Yongjun Xu,
\newblock ``Low-redundancy distillation for continual learning,''
\newblock {\em Pattern Recognition}, p. 111712, 2025.

\bibitem{xu2021privacypreservingmachinelearningmethods}
Runhua Xu, Nathalie Baracaldo, and James Joshi,
\newblock ``Privacy-preserving machine learning: Methods, challenges and directions,'' 2021.

\bibitem{dosovitskiy2021imageworth16x16words}
Alexey Dosovitskiy, Lucas Beyer, Alexander Kolesnikov, Dirk Weissenborn, Xiaohua Zhai, Thomas Unterthiner, Mostafa Dehghani, Matthias Minderer, Georg Heigold, Sylvain Gelly, Jakob Uszkoreit, and Neil Houlsby,
\newblock ``An image is worth 16x16 words: Transformers for image recognition at scale,'' 2021.

\bibitem{li2021prefix}
Xiang~Lisa Li and Percy Liang,
\newblock ``Prefix-tuning: Optimizing continuous prompts for generation,''
\newblock {\em arXiv preprint arXiv:2101.00190}, 2021.

\bibitem{vaswani2017attention}
Ashish Vaswani, Noam Shazeer, Niki Parmar, Jakob Uszkoreit, Llion Jones, Aidan~N Gomez, {\L}ukasz Kaiser, and Illia Polosukhin,
\newblock ``Attention is all you need,''
\newblock {\em Advances in neural information processing systems}, vol. 30, 2017.

\bibitem{zhang2023adapter}
Wentao Zhang, Yujun Huang, Tong Zhang, Qingsong Zou, Wei-Shi Zheng, and Ruixuan Wang,
\newblock ``Adapter learning in pretrained feature extractor for continual learning of diseases,''
\newblock in {\em International Conference on Medical Image Computing and Computer-Assisted Intervention}. Springer, 2023, pp. 68--78.

\bibitem{2009Learning}
A.~Krizhevsky and G.~Hinton,
\newblock ``Learning multiple layers of features from tiny images,''
\newblock {\em Handbook of Systemic Autoimmune Diseases}, vol. 1, no. 4, 2009.

\bibitem{hendrycks2021many}
Dan Hendrycks, Steven Basart, Norman Mu, Saurav Kadavath, Frank Wang, Evan Dorundo, Rahul Desai, Tyler Zhu, Samyak Parajuli, Mike Guo, et~al.,
\newblock ``The many faces of robustness: A critical analysis of out-of-distribution generalization,''
\newblock in {\em Proceedings of the IEEE/CVF international conference on computer vision}, 2021, pp. 8340--8349.

\end{thebibliography}

\end{document}